\documentclass[sigconf]{acmart}

\usepackage{latexsym}
\usepackage{soul}
\usepackage{url}
\usepackage[utf8]{inputenc}
\usepackage{graphicx}
\usepackage{booktabs}
\usepackage{algorithm}
\usepackage{algorithmic}
\usepackage{mathtools}
\usepackage{amsfonts}
\usepackage{blindtext}
\usepackage{amssymb,amsmath,amsthm,enumitem}
\usepackage[font=small,belowskip=-12pt,aboveskip=0pt]{caption}

\newcommand\BibTeX{B\textsc{ib}\TeX}
\newcommand\myeq{\stackrel{\mathclap{\normalfont\mbox{def}}}{=}}

\AtBeginDocument{%
  \providecommand\BibTeX{{%
    \normalfont B\kern-0.5em{\scshape i\kern-0.25em b}\kern-0.8em\TeX}}}

\copyrightyear{2020}
\acmYear{2020}
\setcopyright{iw3c2w3}
\acmConference[WWW '20]{Proceedings of The Web Conference 2020}{April 20--24, 2020}{Taipei, Taiwan}
\acmBooktitle{Proceedings of The Web Conference 2020 (WWW '20), April 20--24, 2020, Taipei, Taiwan}
\acmPrice{}
\acmDOI{10.1145/3366423.3380018}
\acmISBN{978-1-4503-7023-3/20/04}

\begin{document}

\title{Dolphin: A Spoken Language Proficiency Assessment System for Elementary Education}

\author{The Dolphin Team}
\authornote{Authors: Wenbiao Ding, Guowei Xu, Tianqiao Liu, Weiping Fu, Yujia Song, Chaoyou Guo, Cong Kong, Songfan Yang, Gale Yan Huang, Zitao Liu. All authors contributed equally to this research.}
\authornote{Corresponding author: Zitao Liu. Email: liuzitao@100tal.com.}
\affiliation{
  \institution{TAL Education Group, Beijing, China}
}

%

\begin{abstract}
Spoken language proficiency is critically important for children's growth and personal development. Due to the limited and imbalanced educational resources in China, elementary students barely have chances to improve their oral language skills in classes. Verbal fluency tasks (VFTs) were invented to let the students practice their spoken language proficiency after school. VFTs are simple but concrete math related questions that ask students to not only report answers but speak out the entire thinking process. In spite of the great success of VFTs, they bring a heavy grading burden to elementary teachers. To alleviate this problem, we develop Dolphin, a spoken language proficiency assessment system for Chinese elementary education. Dolphin is able to automatically evaluate both phonological fluency and semantic relevance of students' VFT answers. We conduct a wide range of offline and online experiments to demonstrate the effectiveness of Dolphin. In our offline experiments, we show that Dolphin improves both phonological fluency and semantic relevance evaluation performance when compared to state-of-the-art baselines on  real-world educational data sets. In our online A/B experiments, we test Dolphin with 183 teachers from 2 major cities (Hangzhou and Xi'an) in China for 10 weeks and the results show that VFT assignments grading coverage is improved by 22\%. 
\end{abstract}

\ccsdesc[500]{Applied computing~Interactive learning environments}
\ccsdesc[300]{Applied computing~Computer-assisted instruction}
\ccsdesc[300]{Applied computing~E-learning}
\ccsdesc[300]{Computing methodologies~Discourse, dialogue and pragmatics}

\keywords{Spoken language proficiency, Assessment, Verbal fluency, Disfluency detection, K-12 education}


\maketitle




\section{Introduction}
\begin{figure}[!tpbh]
\centering
\includegraphics[width=0.4\textwidth] {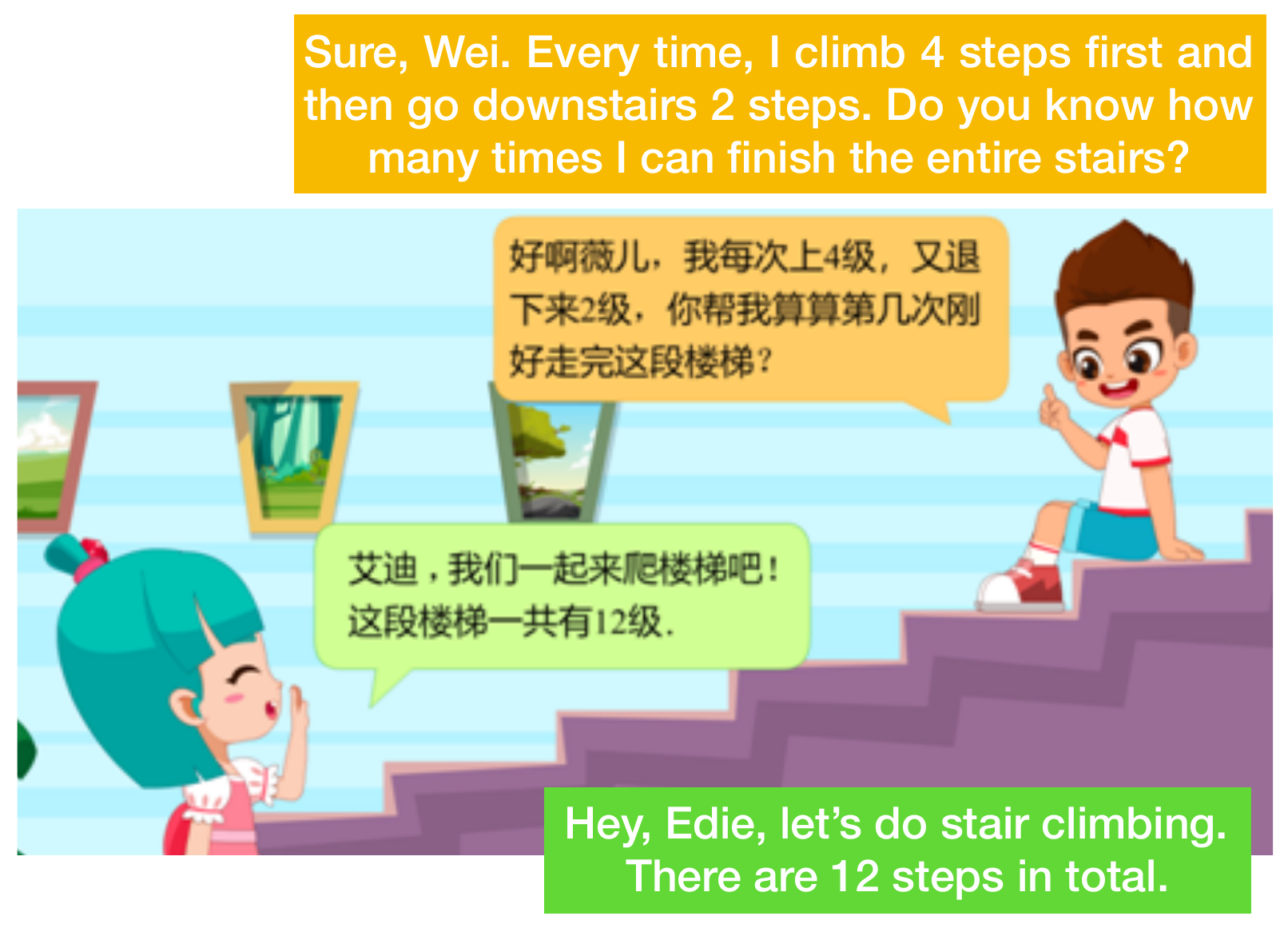}
\caption{One concrete example of our VFTs. It is originally in Chinese and is translated in English for illustration purpose.}
\label{fig:example}
\end{figure}

Spoken language proficiency and verbal fluency usually reflect cognitive flexibility and functions of the brain which are very crucial for personal development \cite{cohen1999verbal,berninger1992gender}. However, developing such skills is very difficult. It requires a large number of oral practices. Studies have shown that it's better to improve and practice verbal fluency during school ages (in elementary school) compared to adult period \cite{baron2009visuospatial,riva2000developmental,gaillard2000functional}. In China, similar to many developing countries, due to the incredibly imbalanced ratio of teachers and elementary students, students barely have chances to speak during a 45-minute class, let alone to practice their verbal fluency abilities. This leads to the fact that elementary school children usually fail to reach satisfactory levels of spoken language proficiency \cite{owens2017classroom,d2015multimodal,olney2017assessing,chen2019multimodal,liu2019automatic}. In order to improve the Chinese elementary students' verbal abilities, we create verbal fluency tasks (VFTs) to help students practice their spoken language proficiency. Figure \ref{fig:example} illustrates one example of our VFTs.

We create VFTs with concrete real-life scenarios. In VFTs we encourage students to ``think aloud": speak out their entire thinking process and not just focus on the answer itself. Teachers first use a mobile App to select VFTs from a VFT bank and assign them to students. Parents will take video clips when students are ``thinking aloud" and the video clips will be automatically collected and graded. The disfluent score not only depends on the occurrence of interregnum words, like \emph{uh}, \emph{I mean}, but also considers long-time silence, out-of-logic sentences, etc. The entire educational design of VFTs is highly recognized by students, parents and educational scholars. However, it brings heavy weekly workloads to teachers when grading every student's VFT clips. In China, an elementary class has 50 students and a teacher holds 4 classes on average. The average length of VFT video clips is about 2 minutes. From our survey, grading VFTs itself brings 10 hours extra workloads to elementary teachers. 

In order to leverage the effectiveness of VFTs and reduce the grading workloads of our elementary teachers, we develop Dolphin, a spoken language proficiency assessment system that is able to automatically evaluate VFTs from both phonological and semantic perspectives. In terms of evaluating phonological fluency, we design a novel deep neural network to learn an effective representation of each video clip and then use logistic regression to predict the phonological fluency score. For detecting semantic relevance, we design a Transformer based multi-attention network to capture semantic relevance between clips and VFTs. We evaluate Dolphin in both offline and online settings. In our offline experiments, we demonstrate the effectiveness of our phonological and semantic relevance scorers by using our real-world educational data sets. In the online A/B experiments, we enable Dolphin for 183 elementary teachers from Hangzhou and Xi'an to help them automatically grade their VFT clips. Various business metrics such as VFT clips grading coverage go up significantly as well as ratings from the teacher satisfaction survey.

\section{Related Work}
Disfluency detection is a well studied NLP task that is informally defined as identifying the interruptions in the normal flow of speech \cite{shriberg1994preliminaries}. In general, disfluency detection methods can be divided into three categories: (1) parsing based approaches \cite{rasooli2013joint,honnibal2014joint}; (2) sequence tagging based approaches \cite{zayats2014multi,ostendorf2013sequential,ferguson2015disfluency,liu2006enriching,schuler2010broad,zayats2016disfluency,hough2015recurrent}; and (3) noisy channel model \cite{lou2018disfluency}. 

Parsing based disfluency detection methods leverage language parsers to identify the syntactic structure of each sentence and use the syntactic structure to infer the disfluency \cite{rasooli2013joint,honnibal2014joint}. Sequence tagging based approaches find fluent or disfluent words within each sentence by modeling word sequences directly and a wide range of models are proposed, such as conditional random fields \cite{zayats2014multi,ostendorf2013sequential,ferguson2015disfluency}, hidden Markov models \cite{liu2006enriching,schuler2010broad}, recurrent neural networks \cite{zayats2016disfluency,hough2015recurrent}, etc. Noisy channel models assume that each disfluent sentence is generated from a well-formed source sentence with some additive noisy words \cite{lou2018disfluency}. They aim to find the \emph{n}-best candidate disfluency analyses for each sentence and eliminate noisy words by using the probability from language models. 

However, several drawbacks prevent existing approaches from evaluating our real-world VFTs. First, all above methods require \emph{clean} and \emph{massive} labeled corpus for training, which is not realistic in our case. In our scenario, the text is generated from students' VFT video clips by using automatic speech recognition (ASR) service. It may contain mistakes from ASR service and can easily fail existing approaches. Furthermore, parsing and sequence tagging based approaches usually need a large amount of annotated data to cover different cases. This would be very expensive to obtain, especially in the specific domain of elementary education. 

\section{The Dolphin Architecture}
\subsection{The Workflow Overview}
\label{sec:workflow}

\begin{figure*}[!tpbh]
\centering
\includegraphics[width=\textwidth] {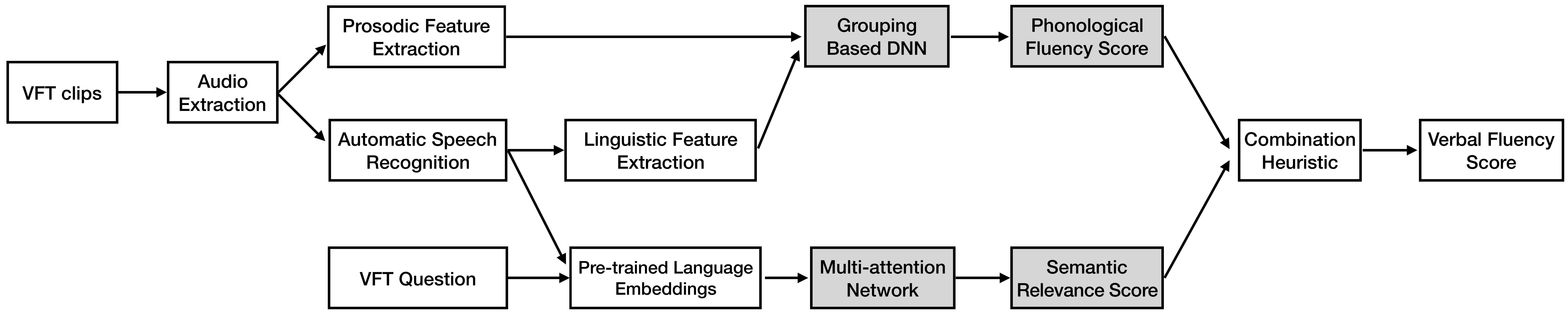}
\caption{The Dolphin workflow. The top/bottom two gray boxes represent the two key components: phonological fluency scorer (Section \ref{sec:phonological}) and semantic relevance scorer (Section \ref{sec:semantic}).}
\label{fig:workflow}
\end{figure*}

The overall workflow of Dolphin is illustrated in Figure \ref{fig:workflow}. The entire workflow is made up of two sub-workflows. In the top sub-workflow, when a VFT clip arrives, we extract its audio track. The audio track is then sent to an open-sourced prosodic feature extractor\footnote{\url{https://www.audeering.com/opensmile/}} and our self-developed automatic speech recognizer (ASR)\footnote{Due to the fact that speech from children are much more different from adults' speech, we trained our own ASR model on a 400-hour children speech recognition data set. The ASR model's word error rate is 86.2\%.}. We extract both speech-related features in the prosodic feature extraction component and linguistic features from ASR output. Then, we combine both  prosodic and linguistic features and utilize a deep neural network (DNN) to learn an effective representation for each clip and predict the phonological fluency by a regularized logistic regression. In the bottom sub-workflow, we first obtain pre-trained language embeddings of both VFT question and ASR transcriptions. After that, we learn a multi-attention network based on transformers to compute semantic relevance scores. Finally we use a combination heuristic to derive verbal fluency scores from phonological fluency and semantic relevance results. We will discuss more details of our phonological fluency scorer and semantic relevance scorer in Section \ref{sec:phonological} and Section \ref{sec:semantic}. The combination heuristic is designed by our teaching professionals based on the company's business logic and we omit its details in this paper.

\subsection{Phonological Fluency Scorer}
\label{sec:phonological}


\subsubsection{Raw Features}
\label{sec:raw}

In order to capture the phonological fluency of each video clip, we extract raw features from both text and audio, which can be summarized into four categories:

\begin{itemize}
\item Word level features, which contain features such as statistics of part-of-speech tags, number of consecutive duplicated words, number of interregnum words \footnote{Interregnum word is an optional part of a disfluent structure that consists of a filled pause \emph{uh} or a discourse marker \emph{I mean}.}, etc.
\item Sentence level features, which contain different statistics from sentence perspective, such as distribution of clip voice length of each sentence, number of characters in each sentence, voice speed of each sentence, etc.
\item Instance (clip) level features, which contain features like total number of characters and sentences, number of long silence that is longer than 5, 3, or 1 seconds, the proportion of effective talking time to clip duration, etc.
\item Prosodic features, which contain speech-related features such as signal energy, loudness, Mel-frequency cepstral coefficients (MFCC), etc.
\end{itemize}

Please note that the ASR not only generates the text transcriptions but the start and end timestamps for each sentence. Such information is very useful for computing features such as voice speed, silence duration percentage, etc. 

\subsubsection{Grouping Based Neural Architecture}

Even though we collect various raw features from many dimensions (discussed in Section \ref{sec:raw}), they tend to be very sparse and directly feeding them into existing predictive models yields poor performance. Similar to \cite{krizhevsky2012imagenet,hinton2012deep,bengio2013representation,xu2019learning}, we utilize a deep neural network to conduct non-linear feature transformation. However, labeled examples are very limited in the real-world scenarios. On the one hand, labeling VFT video clips requires professional teaching knowledge. Thus, the available labeling resource is very limited. On the other hand, labeling one student answer requires the labeler to watch the entire 2-minute clip, which is much more time-consuming compared to standard NLP labeling tasks. With very limited annotated data, many deep representation models may easily run into overfitting problems and become inapplicable.

To address this issue, instead of directly training discriminative representation models from a small amount of annotated labels, we develop a grouping based deep architecture to re-assemble and transform limited labeled examples into many training groups. We include both positive and negative examples into each group. Within each group, we maximize the conditional likelihood of one positive example given another positive example and at the same time, we minimize the conditional likelihood of one positive example given several negative examples. Different from traditional metric learning approaches that focus on learning distance between pairs, our approach aims to generate a more difficult scenario that considers distances between not only positive examples but negative examples.

More specifically, let $(\cdot)^+$ and $(\cdot)^-$ be the indicator of positive and negative examples. Let $\mathcal{D}^+ = \{ \mathbf{x}_i^+ \}_{i=1}^n$ and $\mathcal{D}^- = \{ \mathbf{x}_l^- \}_{l=1}^m$ be the collection of positive and negative example sets, where $n$ and $m$ represent the number of total positive and negative examples in $\mathcal{D}^+$ and $\mathcal{D}^-$. In our grouping based DNN, for each positive example $\mathbf{x}_i^+$, we select another positive example $\mathbf{x}_j^+$ from $\mathcal{D}^+$, where $\mathbf{x}_i^+ \neq \mathbf{x}_j^+$. Then, we randomly select $k$ negative examples from $\mathcal{D}^-$, i.e., $\mathbf{x}_{1}^-, \cdots, \mathbf{x}_{k}^-$. After that, we create a group $\mathbf{g}_s$ by combining the positive pair and the $k$ negative examples, i.e., $\mathbf{g}_s = <\mathbf{x}_i^+, \mathbf{x}_j^+, \mathbf{x}_{1}^-, \cdots, \mathbf{x}_{k}^->$. By using the grouping strategy, we can create $O(n^2 m^k)$ groups for training theoretically. Let $\mathcal{G}$ be the entire collection of  groups, i.e., $\mathcal{G} = \{\mathbf{g}_1, \mathbf{g}_2, \cdots, \mathbf{g}_K\}$ where $K$ is the total number of groups.

\begin{figure}[!tpbh]
\centering
\includegraphics[width=0.475\textwidth] {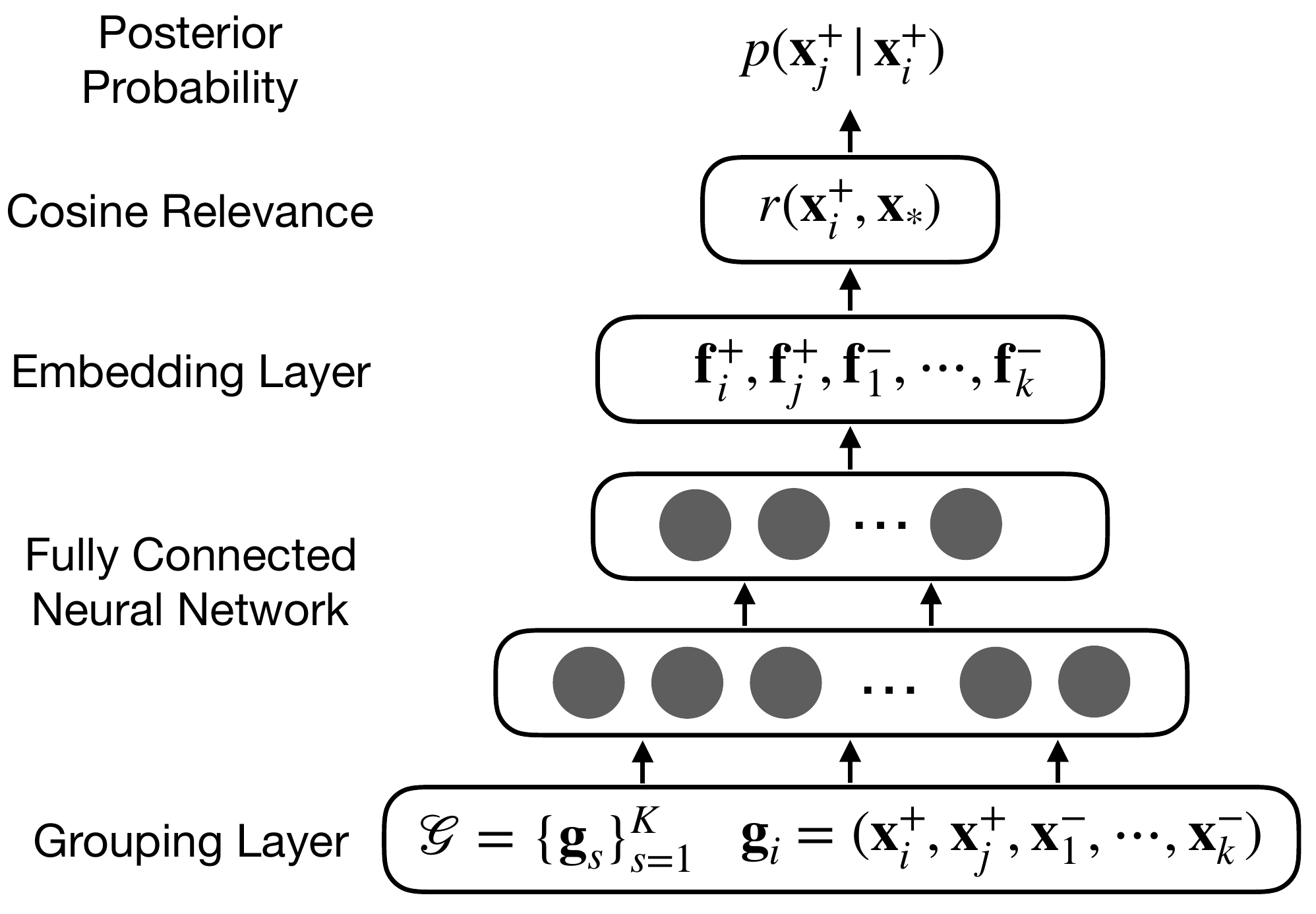}
\caption{The overview of our grouping based deep architecture.}
\label{fig:DSSM_Structure}
\end{figure}

After the grouping procedure, we treat each group $\mathbf{g}_s$ as a training sample and feed $\mathbf{g}_s$s into a standard DNN for learning robust representations. The inputs to the DNN are raw features extracted from Section \ref{sec:raw} and the output of the DNN is a low-dimensional representation feature vector. Inside the DNN, we use the multi-layer fully-connected non-linear projections to learn compact representations as shown in Figure \ref{fig:DSSM_Structure}. 

\noindent \textbf{Model Learning} Inspired by the discriminative training approaches in information retrieval, we propose a supervised training approach to learn model parameters by maximizing the conditional likelihood of retrieving positive example $\mathbf{x}_j^+$ given positive example $\mathbf{x}_i^+$ from group $\mathbf{g}_s$. More formally, let $\mathbf{f}_*$ be the learned representation of $\mathbf{x}_*$ from DNN, where $\mathbf{x}_* \in \mathbf{g}_s$. Similarly, $\mathbf{f}_i^+$ and $\mathbf{f}_l^-$ represent the embeddings of the positive example $\mathbf{x}_i^+$ and negative example $\mathbf{x}_l^-$. Then, the relevance score in the representation space within a group is then measured as $\mbox{cosine}(\mathbf{f}_i^+, \mathbf{f}_*)$, i.e., $r(\mathbf{x}_i^+, \mathbf{x}_*) \myeq  \mbox{cosine}(\mathbf{f}_i^+, \mathbf{f}_*)$.

In our representation learning framework, we compute the posterior probability of $\mathbf{x}_j^+$ in group $\mathbf{g}_s$ given $\mathbf{x}_i^+$ from the cosine relevance score between them through a softmax function

\begin{equation}
\label{eq:softmax}
\nonumber
p(\mathbf{x}_j^+|\mathbf{x}_i^+) = \frac{\exp \big(\eta \cdot r(\mathbf{x}_i^+, \mathbf{x}_j^+)\big)}{\sum_{\mathbf{x}_* \in \mathbf{g}_s, \mathbf{x}_* \neq \mathbf{x}_i^+} \exp \big(\eta \cdot r(\mathbf{x}_i^+, \mathbf{x}_*)\big) }
\end{equation}

\noindent where $\eta$ is a smoothing hyper parameter in the softmax function, which is set empirically on a held-out data set in our experiment. 

Hence, given a collection of groups $\mathcal{G}$, we optimized the DNN model parameters by maximizing the sum of log conditional likelihood of finding a positive example $\mathbf{x}_j^+$ given the paired positive example $\mathbf{x}_i^+$ from group $\mathbf{g}_s$, i.e., $\mathcal{L}(\Omega) = - \sum \log p(\mathbf{x}_j^+|\mathbf{x}_i^+)$, where $\Omega$ is the parameter set of the DNN. Since $\mathcal{L}(\Omega)$ is differentiable with respect to $\Omega$, we use gradient based optimization approaches to train the DNN.

\subsection{Semantic Relevance Scorer}
\label{sec:semantic}

\begin{figure*}[!tpbh]
\centering
\includegraphics[width=\textwidth]{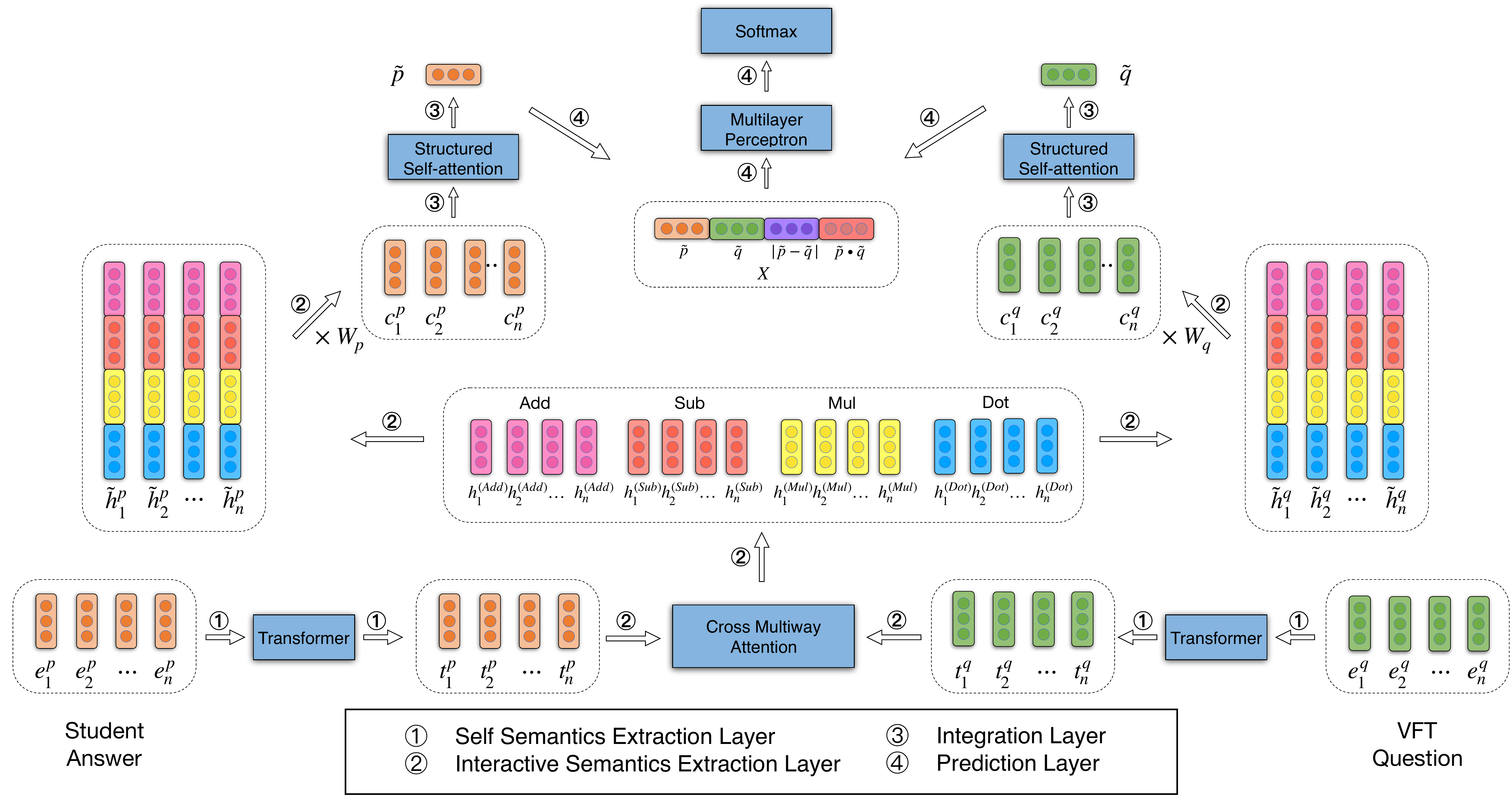}
\caption{The overview of our model. We use colors to better represent intermediate manipulations of hidden states or embeddings.}
\label{fig:semantic_model}
\end{figure*}

Semantic analysis of video clips is also very important for our spoken language proficiency assessment. After a detailed study on teachers' manually grading cases, we observe that some naughty students fool our phonological fluency scorer by talking about completely irrelevant topics, such as answering different questions, casual life recording, or even playing video games. Therefore, we develop an end-to-end predictive framework to evaluate the semantic relationship between the students' video clips and the VFT questions. 

Our framework is based on the recently developed Transformer architecture \cite{vaswani2017attention}. It dispenses with any recurrence and convolutions entirely and is based solely on attention mechanisms. Let $\mathcal{P} = \{\mathbf{e}^p_i\}_{i=1}^n$ be the word embeddings of ASR transcriptions from student's video clip and $\mathcal{Q} = \{\mathbf{e}^q_j\}_{j=1}^n$ be the word embeddings of the assigned VFT question\footnote{Chinese Word Embedding from Tencent AI Lab is used in the following experiments, see \url{https://ai.tencent.com/ailab/nlp/embedding.html}.}. $\mathbf{e}^p_i$ and $\mathbf{e}^q_j$ are the $k$-dimensional pre-trained word embeddings. Without the loss of generality, we convert sentences in both $\mathcal{P}$ and $\mathcal{Q}$ to $n$ words. The proposed framework is shown in Figure \ref{fig:semantic_model} and it is built upon four crucial layers, which are described in the following sections.

\subsubsection{Self Semantics Extraction Layer}
\label{sec:self}

As shown in Figure \ref{fig:semantic_model}, we adopt a Transformer encoder to encode the student answer $\mathcal{P}$ and the VFT question $\mathcal{Q}$ as their contextual representations, i.e., $\{\mathbf{t}^p_i\}_{i=1}^n = {\rm Transformer}(\mathcal{P})$ and $\{\mathbf{t}^q_j\}_{j=1}^n = {\rm Transformer}(\mathcal{Q})$, respectively. The Transformer encoder consists of multiple identical blocks which contain multi-head attention sub-layers and position-wise fully connected feed-forward sub-layers \cite{vaswani2017attention}. Through the multi-head self-attention mechanism, the model captures the long-range dependencies among the words in the original texts and re-encodes each word as a combination of all the words related to it. After that, we get the contextual embeddings of answers, i.e., $\{\mathbf{t}_i^p\}_{i=1}^n$, and VFT questions, i.e., $\{\mathbf{t}_j^q\}_{j=1}^n$. 

\subsubsection{Interactive Semantics Extraction Layer}

Following previous work \cite{rocktaschel2015reasoning,wang2016sentence,tan2018multiway}, we use a cross multiway attention mechanism to extract the semantic relation between the student answers and the VFT questions. Specifically, we attend contextual embeddings of student answers to VFT questions and attend VFT questions to student answers as well. We use additive (\emph{Add}), subtractive (\emph{Sub}), multiplicative (\emph{Mul}), dot-product (\emph{Dot}) functions as our cross-attention functions to compute four types of attentions scores that are used to obtain the corresponding weighted-sum representations. Let $\odot$ be the operation of element-wise dot product. Let $s_{ij}^{(*)}$, $a_{ij,p}^{(*)}$, $a_{ij,q}^{(*)}$ be the raw and normalized attention scores from different attention functions  and let $\mathbf{h}_{i,p}^{(*)}$ $\mathbf{h}_{j,q}^{(*)}$ be the representations weighted by different types of attention scores where $*$ indicates different attention functions and it can be Add, Sub, or Dot. The detailed descriptions of different attention functions and representations are shown in Table \ref{tab:attention}. In Table \ref{tab:attention}, $\mathbf{v}^{({\rm Add})} \in \mathbb{R}^k$, $\mathbf{v}^{({\rm Sub})} \in \mathbb{R}^k$, $\mathbf{v}^{({\rm Dot})} \in \mathbb{R}^k$, $\mathbf{W}^{({\rm Mul})} \in \mathbb{R}^{k \times k}$, $\mathbf{W}^{({\rm Dot})} \in \mathbb{R}^{k \times k}$, $\mathbf{W}^{({\rm Add})}_p \in \mathbb{R}^{k \times k}$, $\mathbf{W}^{({\rm Sub})}_p \in \mathbb{R}^{k \times k}$ and $\mathbf{W}^{({\rm Add})}_q \in \mathbb{R}^{k \times k}$, $\mathbf{W}^{({\rm Sub})}_q \in \mathbb{R}^{k \times k}$ are learned parameters in our framework. ${\rm softmax}_p$ and ${\rm softmax}_q$ are softmax functions of normalizing attention scores for both student answers and VFT questions. The final representations of student answers, i.e., $\{\mathbf{c}^p_i\}_{i=1}^n$ and VFT questions, i.e., $\{\mathbf{c}^q_j\}_{j=1}^n$ are computed by a linear transformation on the concatenation of four different attention-weighted embeddings, i.e.,

\begin{table*}[!t]
\centering
\footnotesize
\caption{The descriptions of different attention functions. Add, Sub, Mul and Dot represent the additive, subtractive, multiplicative and dot-product attention functions.}
\label{tab:attention}
\begin{tabular}{l|l|l|l|l|l}
\toprule
 & \textbf{Attentive Function} & \textbf{Attention Score (Answer)} & \textbf{Answer Embedding} & \textbf{Attention Score (Question)} & \textbf{Question Embedding} \\ \midrule

\textbf{Add} & $s_{ij}^{({\rm Add})} = {\mathbf{v}^{({\rm Add})}}^\top {\rm tanh}(\mathbf{W}_p^{({\rm Add})}\mathbf{t}_i^p+\mathbf{W}_q^{({\rm Add})}\mathbf{t}_j^q)$ & $a^{({\rm Add})}_{ij,p} = {\rm softmax}_p(s_{ij}^{({\rm Add})})$ & $\mathbf{h}_{i,p}^{({\rm Add})} = \sum_{j=1}^{n} a^{({\rm Add})}_{ij,p} \mathbf{t}_j^q$ & $a^{({\rm Add})}_{ij,q} = {\rm softmax}_q(s_{ij}^{({\rm Add})})$ & $\mathbf{h}_{j,q}^{({\rm Add})} = \sum_{i=1}^{n} a^{({\rm Add})}_{ij,q} \mathbf{t}_i^p$ \\ \midrule

\textbf{Sub} & $s_{ij}^{({\rm Sub})} = {\mathbf{v}^{({\rm Sub})}}^\top {\rm tanh}(\mathbf{W}_p^{({\rm Sub})}\mathbf{t}_i^p-\mathbf{W}_q^{({\rm Sub})}\mathbf{t}_j^q)$ & $a^{({\rm Sub})}_{ij,p} = {\rm softmax}_p(s_{ij}^{({\rm Sub})})$ & $\mathbf{h}_{i,p}^{({\rm Sub})} = \sum_{j=1}^{n} a^{({\rm Sub})}_{ij,p} \mathbf{t}_j^q$ & $a^{({\rm Sub})}_{ij,q} = {\rm softmax}_q(s_{ij}^{({\rm Sub})})$ & $\mathbf{h}_{j,q}^{({\rm Sub})} = \sum_{i=1}^{n} a^{({\rm Sub})}_{ij,q} \mathbf{t}_i^p$ \\ \midrule

\textbf{Mul} & $s_{ij}^{({\rm Mul})} = {\mathbf{t}_i^p}^\top \mathbf{W}^{({\rm Mul})} \mathbf{t}_j^q$ & $a^{({\rm Mul})}_{ij,p} = {\rm softmax}_p(s_{ij}^{({\rm Mul})})$ & $\mathbf{h}_{i,p}^{({\rm Mul})} = \sum_{j=1}^{n} a^{({\rm Mul})}_{ij,p} \mathbf{t}_j^q$ & $a^{({\rm Mul})}_{ij,q} = {\rm softmax}_q(s_{ij}^{({\rm Mul})})$ & $\mathbf{h}_{j,q}^{({\rm Mul})} = \sum_{i=1}^{n} a^{({\rm Mul})}_{ij,q} \mathbf{t}_i^p$ \\ \midrule

\textbf{Dot} & $s_{ij}^{({\rm Dot})} = {\mathbf{v}^{({\rm Dot})}}^\top {\rm tanh}(\mathbf{W}^{({\rm Dot})}(\mathbf{t}_i^p \odot \mathbf{t}_j^q))$ & $a^{({\rm Dot})}_{ij,p} = {\rm softmax}_p(s_{ij}^{({\rm Dot})})$ & $\mathbf{h}_{i,p}^{({\rm Dot})} = \sum_{j=1}^{n} a^{({\rm Dot})}_{ij,p} \mathbf{t}_j^q$ & $a^{({\rm Dot})}_{ij,q} = {\rm softmax}_q(s_{ij}^{({\rm Dot})})$ & $\mathbf{h}_{j,q}^{({\rm Dot})} = \sum_{i=1}^{n} a^{({\rm Dot})}_{ij,q} \mathbf{t}_i^p$  \\  
\bottomrule
\end{tabular}
\end{table*}

\begin{align*}
    \mathbf{c}_i^p & = \mathbf{W}_p \Tilde{\mathbf{h}}^p_i=\mathbf{W}_p[\mathbf{h}_{i,p}^{({\rm Add})}:\mathbf{h}_{i,p}^{({\rm Sub})}:\mathbf{h}_{i,p}^{({\rm Mul})}:\mathbf{h}_{i,p}^{({\rm Dot})}] \\
    \mathbf{c}_j^q & = \mathbf{W}_q \Tilde{\mathbf{h}}^q_j=\mathbf{W}_q[\mathbf{h}_{j,q}^{({\rm Add})}:\mathbf{h}_{j,q}^{({\rm Sub})}:\mathbf{h}_{j,q}^{({\rm Mul})}:\mathbf{h}_{j,q}^{({\rm Dot})}] \\
\end{align*}

\noindent where $[a:b]$ denotes the concatenation of vector $a$ and $b$ and $\mathbf{W}_p \in \mathbb{R}^{k \times 4k}$ and $\mathbf{W}_q \in \mathbb{R}^{k \times 4k}$ are the linear projection matrices.

\subsubsection{Integration Layer}
\label{sec:integration}

We first stack $\{\mathbf{c}^p_i\}_{i=1}^n$ and $\{\mathbf{c}^q_j\}_{j=1}^n$ into matrix form $\mathbf{C}_p$ and $\mathbf{C}_q$, i.e., $\mathbf{C}_p \in \mathbb{R}^{k \times n}, \mathbf{C}_p = [\mathbf{c}^p_1; \mathbf{c}^p_2; \cdots; \mathbf{c}^p_n]$ and $\mathbf{C}_q \in \mathbb{R}^{k \times n}, \mathbf{C}_q = [\mathbf{c}^q_1; \mathbf{c}^q_2; \cdots; \mathbf{c}^q_n]$ and then convert the aggregated representation $\mathbf{C}_p$ and $\mathbf{C}_q$ of all positions into a fixed-length vector with structured self-attention pooling \cite{lin2017structured}, which is defined as

\begin{align*}
\Tilde{\mathbf{p}} & = \mathbf{C}_p{\rm softmax}(\mathbf{w}_{p1}^\top {\rm tanh}(\mathbf{W}_{p2}{\mathbf{C}_p})) \\
\Tilde{\mathbf{q}} & = \mathbf{C}_q{\rm softmax}(\mathbf{w}_{q1}^\top {\rm tanh}(\mathbf{W}_{q2}{\mathbf{C}_q})) 
\end{align*}

\noindent where $\mathbf{w}_{p1} \in \mathbb{R}^{k}$, $\mathbf{w}_{q1} \in \mathbb{R}^{k}$, $\mathbf{W}_{p2} \in \mathbb{R}^{k \times k}$ and $\mathbf{W}_{q2} \in \mathbb{R}^{k \times k}$ are the learned parameters.

\subsubsection{Prediction Layer}
\label{sec:prediction}

Similar to \cite{conneau2017supervised}, we use three matching methods to extract relations between the student answer and the reference answer: (1) concatenation of the two representations ($\Tilde{\mathbf{p}}$, $\Tilde{\mathbf{q}}$); (2) element-wise product $\Tilde{\mathbf{p}} * \Tilde{\mathbf{q}}$; and
(3) absolute element-wise difference |$\Tilde{\mathbf{p}} - \Tilde{\mathbf{q}}$|, i.e., $\mathbf{x} = [\Tilde{\mathbf{p}}:\Tilde{\mathbf{q}}:|\Tilde{\mathbf{p}}-\Tilde{\mathbf{q}}|:\Tilde{\mathbf{p}}\cdot\Tilde{\mathbf{q}}]$. The resulting vector $\mathbf{x}$, which captures information from both the student answer and the reference answer, is fed into a multilayer perceptron classifier to output the probability (score) of the semantic analysis tasks. The objective function is to minimize the cross entropy of the relevance labels.

\section{Experiments}
\subsection{Phonological Fluency Experiment}
\label{sec:phonological_exp}

In this experiment, we collect 880 VFT video clips from students in 2nd grade. In each clip, a student talks about his or her entire thinking process of solving a math question. Our task is to score the fluency of the student's entire speech on a scale between 0 and 1. To obtain a robust annotated label, each video clip is annotated as either fluent or disfluent by 11 teaching professionals and we use the majority voted labels as our ground truth. The proportion of positive (fluent) examples is 0.60, the Kappa coefficient is 0.65 in this data set. 

We select several recent work on embedding learning from limited data as our baselines. More specifically, we have 

\begin{itemize}

\item Logistic regression, i.e., \emph{LR}. We train an LR classifier on raw features (discussed in Section \ref{sec:raw}); 

\item Siamese networks, i.e., \emph{SiameseNet} \cite{koch2015siamese}. We train a Siamese network that takes a pair of examples and trains the embeddings so that the distance between them is minimized if they're from the same class and is greater than some margin value if they represent different classes; 

\item Triplet networks, i.e., \emph{TripleNet} \cite{schroff2015facenet}. We train a triplet network that takes an anchor, a positive (of same class as an anchor) and negative (of different class than an anchor) examples. The objective is to learn embeddings such that the anchor is closer to the positive example than it is to the negative example by some margin value; 

\item Relation network for few-shot learning, i.e., \emph{RelationNet} \cite{yang2018learning}. The RelationNet learns to learn a deep distance metric to compare a small number of images within episodes. 

\end{itemize}

Following the tradition to assess representation learning algorithms \cite{bengio2013representation}, we evaluate the classification performance via accuracy and F1 score. We choose logistic regression as the basic classifier. For each task, we conduct a 5-fold cross validation on the data sets and report the average performance in Table \ref{tab:phonological_results}. We conduct pair wise t-tests between our approach and every baseline mentioned above and the results show that the prediction outcomes are statistically significant at 0.05 level. As we can see from Table \ref{tab:phonological_results}, our grouping based approach demonstrates the best performance. This is due to the fact that our discriminative embedding learning directly optimizes the distance between positive and negative examples. By generating all the groups, we want to separate each positive pair with a few negative examples as far as possible. In terms of TripleNet and RelationNet, both of them optimize the distance function between positive and negative indirectly. For example,  TripletNet generates more training samples by creating triplet, which contains anchor, positive and negative examples. It fully utilizes the anchor to connect the positive and negative examples during the training. However, selecting triplet is not straightforward. SiameseNet has the worst performance because it is designed for one-shot learning, which aims to generalize the predictive power of the network not just to new data, but to entirely new classes from unknown distributions. This may lead to the inferior performance in our situation that both positive and negative examples are very limited. 

\begin{table}[!tbpb]
\centering
\small
\caption{\label{tab:phonological_results}Experimental results on phonological relevance detection.}
\begin{tabular}{l|c|c|c|c|c}
\toprule
  & \textbf{LR} & \textbf{SiameseNet} & \textbf{TripleNet} & \textbf{RelationNet} & \textbf{Our} \\ \midrule
\textbf{Accuracy}	& 0.820 & 0.802 & 0.847  &  0.843 & 0.871 \\ \midrule
\textbf{F1 score}	& 0.873 & 0.859 & 0.889  &  0.890 & 0.901 \\ \bottomrule
\end{tabular}
\end{table}

\subsection{Semantic Relevance Experiment}

We collect 120,000 pairs of student's video clip and VFT question for training and evaluating the semantic relevance scorer. It contains 90,000 positive and 30,000 negative pairs, which are annotated by professional annotators. The positive pairs indicate that the video clip and VFT question are semantically relevant. We randomly select 30,000 samples as our test data and use the rest for validation and training. 

We compare our model with several state-of-the-art baselines of sentence similarity and classification to demonstrate the performance of our multi-attention network. More specifically, we choose 

\begin{itemize}

\item Logistic regression, i.e., \emph{LR}. Similar to \cite{arora2016simple}, we compute embeddings of $\mathcal{Q}$ and $\mathcal{P}$ by averaging each word's embedding from a pre-trained look-up table. Then we concatenate the two embedding of $\mathcal{P}$ and $\mathcal{Q}$ as the input for logistic regression;

\item Gradient boosted decision tree, i.e., \emph{GBDT}. We use the same input as LR but train the GBDT model instead; 

\item Multichannel convolutional neural networks, i.e., \emph{TextCNN}. Similar to \cite{kim2014convolutional}, we train  embedding by using CNN on top of pre-trained word vectors; 

\item Sentence encoding by transformer , i.e., \emph{Transformer}. As discussed in \cite{vaswani2017attention}, we use transformer as the encoding module and apply a feed-forward network for classification.

\end{itemize}

Similar to experiments in Section \ref{sec:phonological_exp}, we report the classification accuracy and F1 score in Table \ref{tab:semantic_results}. We conduct pair wise t-tests between our approach and every baseline mentioned above and the results show that the prediction outcomes are statistically significant at 0.05 level. As we can see, our Transformer based multiway attention network shows superior performance among all baselines. It not only aggregates sentence information within Transformer encoder layer, but matches words in both video clips and VFT questions from multiway attention layer. We aggregate all available semantic information by an inside aggregation layer. Due to the lack of any contextual information, LR have the worst classification performance. TextCNN and Transformer improve the accuracy by considering the sentence level sequential information. However, it ignores the semantic relationship between video clips and VFT questions. 

\begin{table}[!thbp]
\centering
\small
\caption{\label{tab:semantic_results}Experimental results on semantic relevance detection.}
\begin{tabular}{l|c|c|c|c|c}
\toprule
  & \textbf{LR} & \textbf{GBDT} & \textbf{TextCNN} & \textbf{Transformer} & \textbf{Our} \\ \midrule
\textbf{Accuracy} & 0.830 & 0.863 & 0.877  &  0.881 &  0.888 \\ \midrule
\textbf{F1 score} & 0.838 & 0.872 & 0.888 & 0.890 & 0.896 \\ \bottomrule
\end{tabular}
\end{table}

\subsection{Online A/B Experiments}

In order to fully demonstrate the value of the Dolphin system, we also conduct the online A/B experiments. In our Dolphin system, automatic grades are given to teachers and teachers have the option to switch to manual grading model. We enable Dolphin for 183 elementary teachers from Hangzhou and Xi'an from 2018.09.05 to 2018.11.19. The total number of weekly graded VFT assignments is around 80,000. In summary, 92\% of teachers enable the Dolphin automatic grading functionality. Among these enabled subjects, 87\% of the automatic gradings of VFT assignments are accepted. Here, we would like to mention that there are 13\% teachers choose to grade manually doesn't mean our grading error rate is 13\%. Furthermore, we compare with teachers that have no access to Dolphin in terms of the overall grading coverage. The results show that Dolphin improves VFT assignments grading coverage\footnote{Coverage is calculated as number of graded assignments divided by total number of assignments.} by 22\%.

\begin{figure}[!tpbh]
\centering
\includegraphics[width=0.5\textwidth] {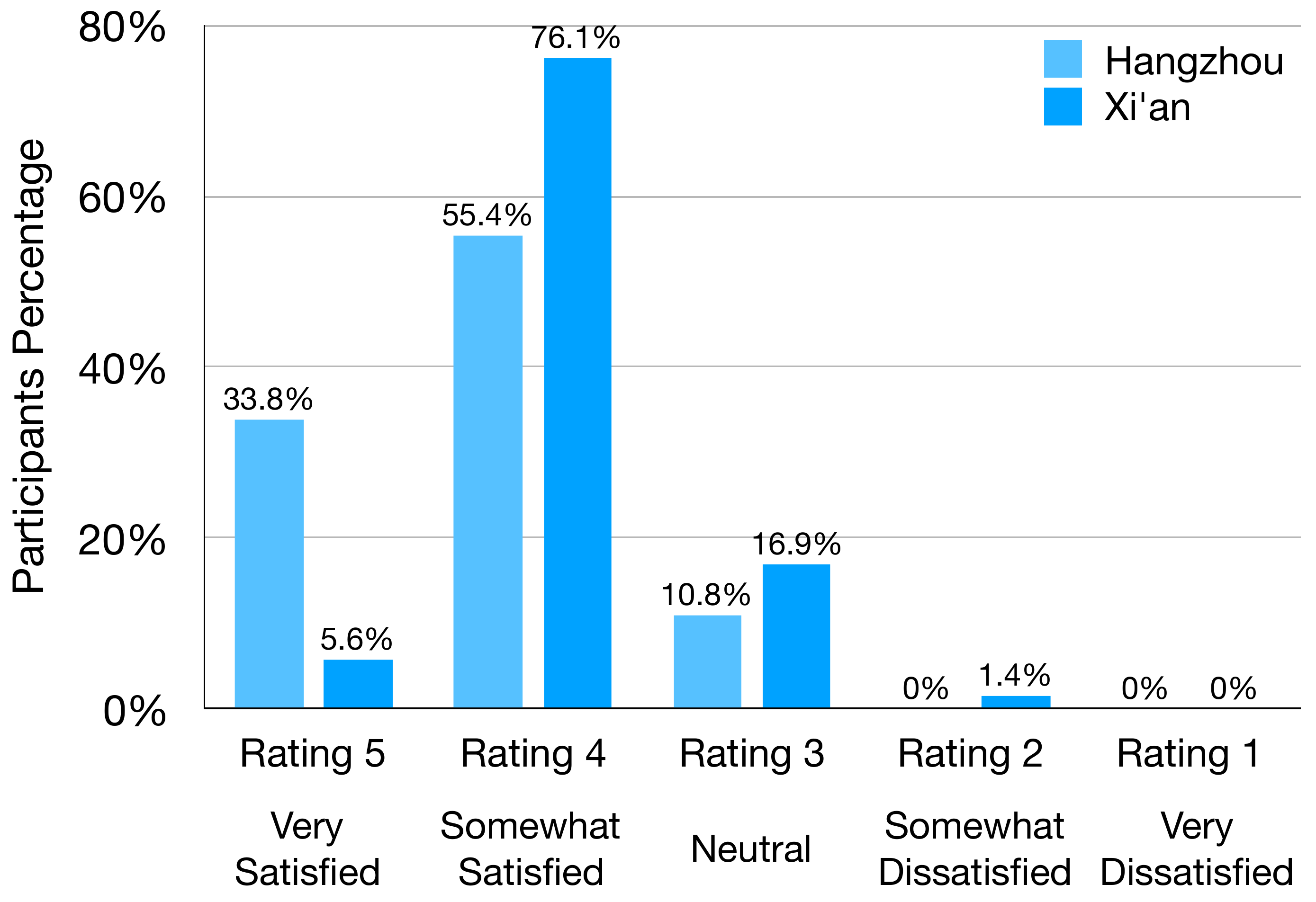}
\caption{Satisfaction rating distributions from teachers participated in our A/B experiments.}
\label{fig:rating}
\end{figure}

We also conduct a questionnaire-based survey about teachers' Dolphin usage experience specifically. The teachers participated in our A/B experiments are asked to give a satisfaction rating from 1 to 5 (5 = Very Satisfied and 1 = Very Dissatisfied). There are 183 teachers participating this survey (85 from Hangzhou and 98 from Xi'an). The satisfaction rating distributions are listed in Figure \ref{fig:rating}. As we can see from Figure \ref{fig:rating}, teachers are pretty positive on this automatic VFT grading tool and the overall dissatisfied ratio (sum of rating 2 and rating 1) is very low (Hangzhou: 0\% and Xi'an: 1.4\%). The very satisfied percentage in Xi'an is much lower than Hangzhou. This is because the VFT assignments were recently introduced to schools in Xi'an and teachers are very serious and conservative to both the VFT assignments and the automatic grading tool. 

\subsection{Discussion of Ethical Issues}

Automatic grading tools are double-edged swords. On the one hand, They greatly reduce the burdens of tedious and repetitive grading workloads for teachers and let them have enough time to prepare class materials and teaching content. On the other hand, it could mislead students and sweep students' problems under the carpet. As a rule of thumb, only use them for skill practice instead of entrance examinations. Moreover, we design policies and heuristics to mitigate the misleading risks. First, we ask both the teachers and parents to randomly check one result in every 5 VFT assignments. They are able to report the wrong grading immediately from the mobile application. Second, when the grading tool gives 3 consecutive low ratings to a child, an alarm will be fired to the corresponding teacher and the teacher is required to check all recent ratings.

\section{Conclusion}
In this paper we present Dolphin, a spoken language proficiency assessment system for elementary students. Dolphin provides more opportunities for Chinese elementary students to practice and improve their oral language skills and at the same time reduces teachers' grading burden. Experiment results in both offline and online environments demonstrate the effectiveness of Dolphin in terms of model accuracy, system usage, users (teacher) satisfaction rating and many other metrics. 


\bibliographystyle{ACM-Reference-Format}
\bibliography{www2020}

\end{document}